\documentclass{article}
\usepackage{arxiv}

\usepackage{cite}
\usepackage{amsmath,amssymb,amsfonts}
\usepackage{algorithmic}
\usepackage{graphicx}
\usepackage{textcomp}

\usepackage{verbatim} 
\usepackage{soul} 
\usepackage{wasysym} 
\usepackage{todonotes}
\usepackage[T1]{fontenc}
\usepackage[utf8]{inputenc}

\title{Sentiment Analysis on Electricity Twitter Posts}

\author{
Pardeep Kaur, Maryam Edalati 
\\
 \textit{Department of Computer Science} \\
    \textit{Norwegian University of Science and Technology}\\
    Gjøvik, Norway \\
    {pardeeka@stud.ntnu.no,  maryame@stud.ntnu.no}
}

\begin{document}
\maketitle

\begin{abstract}
In today’s world, everyone is expressive in some way, and the focus of this project is on people’s opinions about rising electricity prices in United Kingdom and India using data from Twitter, a micro-blogging platform on which people post messages, known as tweets. Because many people's incomes are not good and they have to pay so many taxes and bills, maintaining a home has become a disputed issue these days. Despite the fact that Government offered subsidy schemes to compensate people electricity bills but it is not welcomed by people. In this project, the aim is to perform sentiment analysis on people's expressions and opinions expressed on Twitter. In order to grasp the electricity prices opinion, it is necessary to carry out sentiment analysis for the government and consumers in energy market. Furthermore, text present on these medias are unstructured in nature, so to process them we firstly need to pre-process the data. There are so many feature extraction techniques such as Bag of Words, TF-IDF (Term Frequency-Inverse Document Frequency), word embedding, NLP based features like word count. In this project, we analysed the impact of feature TF-IDF word level on electricity bills dataset of sentiment analysis. We found that by using TF-IDF word level performance of sentiment analysis is 3-4 higher than using N-gram features. Analysis is done using four classification algorithms including Naive Bayes, Decision Tree, Random Forest, and Logistic Regression and considering F-Score, Accuracy, Precision, and Recall performance parameters.
\end{abstract}

\keywords{
Sentiment Analysis \and Machine Learning \and Electricity \and opinion mining \and polarity assessment 
}

\section{Introduction}
\label{sec:intro}

Due to increase in contents over social media such as Twitter, Facebook and Trip advisor, expressing opinions about products, services and government policies among others has increased. Twitter has 336 million active users monthly and it is now a main source of feedback for government, private organizations and other service providers \cite{ahuja2019impact}. On Twitter around 500 millions tweets are produced per day with people sharing their joys and problems in their daily lives \cite{ahuja2019impact}. Twitter, a social media platform, has received a lot of attention from researchers in the recent times \cite{pagolu2016sentiment,batra2021evaluating, stamm2022does, imran2020cross}.

Opinion mining and sentiment analysis domain has widely been studied extensively in many other fields including education \cite{10.1145,sadriu2022automated,edalati2021potential, sandra2022university, kastrati2021sentiment, stamm2022does, altrabsheh2013sa, kastrati2020weakly}, movies\cite{ali2019sentiment, yasen2019movies, salmony2022bert, jain2013prediction}, stock prediction \cite{nguyen2015sentiment, feldman2011stock, bapat2014comprehensive}, product reviews \cite{fang2015sentiment, bibi2022novel, zhao2017weakly}, tracking and understanding people's reactions on pandemics on social media \cite{Fu:2016,electronics10101133, nezhad2022twitter}, etc. Many have started pivoting to social media to analyse emotions and collecting large twitter datasets \cite{batra2021large}. Social media platforms play an essential role during the extreme crisis as individuals use these communication channels to share ideas, opinions, and reactions with others to cope with and react to crises \cite{batra2021evaluating}.   The most common concern on Twitter these days is a rise in energy prices, such as electricity prices. Sharp spikes in electricity prices in recent times have been causing hardship for many households and businesses around the world and risk becoming a driver of social and political tensions,” said IEA Executive Director Fatih Birol. There are more people that are trying to figure out what to do as far as maintaining their household and providing for their families, and so it’s really hard\footnote{https://www.nbcnews.com/news/us-news/-lot-people-need-help-soaring-electric-bills-leave-struggling-pay-rcna16347}. With two thirds (66 percent) of adults in Britain reporting their cost of living increased in the past month, rising energy prices are a growing factor in the squeeze on household budgets\footnote{https://www.bbc.com/news/business-60943192}.

Thousands of people have taken to the streets in the UK in protest against the sharp rise in energy prices, as a cabinet minister said the government could not “completely nullify” the increases. As a result, in this project, we will investigate collective reactions to events expressed on social media. Because of its widespread popularity and ease of access via the API, a special emphasis will be placed on analyzing people’s reactions to increases in electricity bills expressed on Twitter’s social media platform. To that end, tweets from thousands of Twitter users were collected this year after a sharp increase in energy prices in developed and developing countries to understand how different cultures were reacting and responding to this situation.

The core contributions of this study are as following:
\begin{enumerate}
    \item {Collecting a dataset composed of nearly 10000 tweets related to energy prices using Twitter API for academic research and the Python programming language.}
    \item {Selecting two different countries with various incomes, one developing country and one developed country.}
    \item Using a lexicon-based sentiment analysis approach, i.e. TextBlob or Vader to calculate sentiment score and infer a sentiment polarity to each tweet (Positive, Neutral and Negative).
    \item Applying and comparing different machine and deep learning algorithms on the collected and curated dataset.

\end{enumerate}

\section{Related Work}
\label{sec:relatedWork}

Gas, coal and electricity prices have in recent weeks risen to their highest levels in decades \cite{alvarez2021behind}. The strong increases in natural gas prices have prompted substantial switching to the use of coal rather than natural gas to generate electricity in key markets, including the United States, Europe and Asia. The increased use of coal is in turn is driving up CO2 emissions from electricity generation globally \cite{alvarez2021behind}.

\begin{figure}[ht]
    \centering
    \includegraphics[width=15cm]{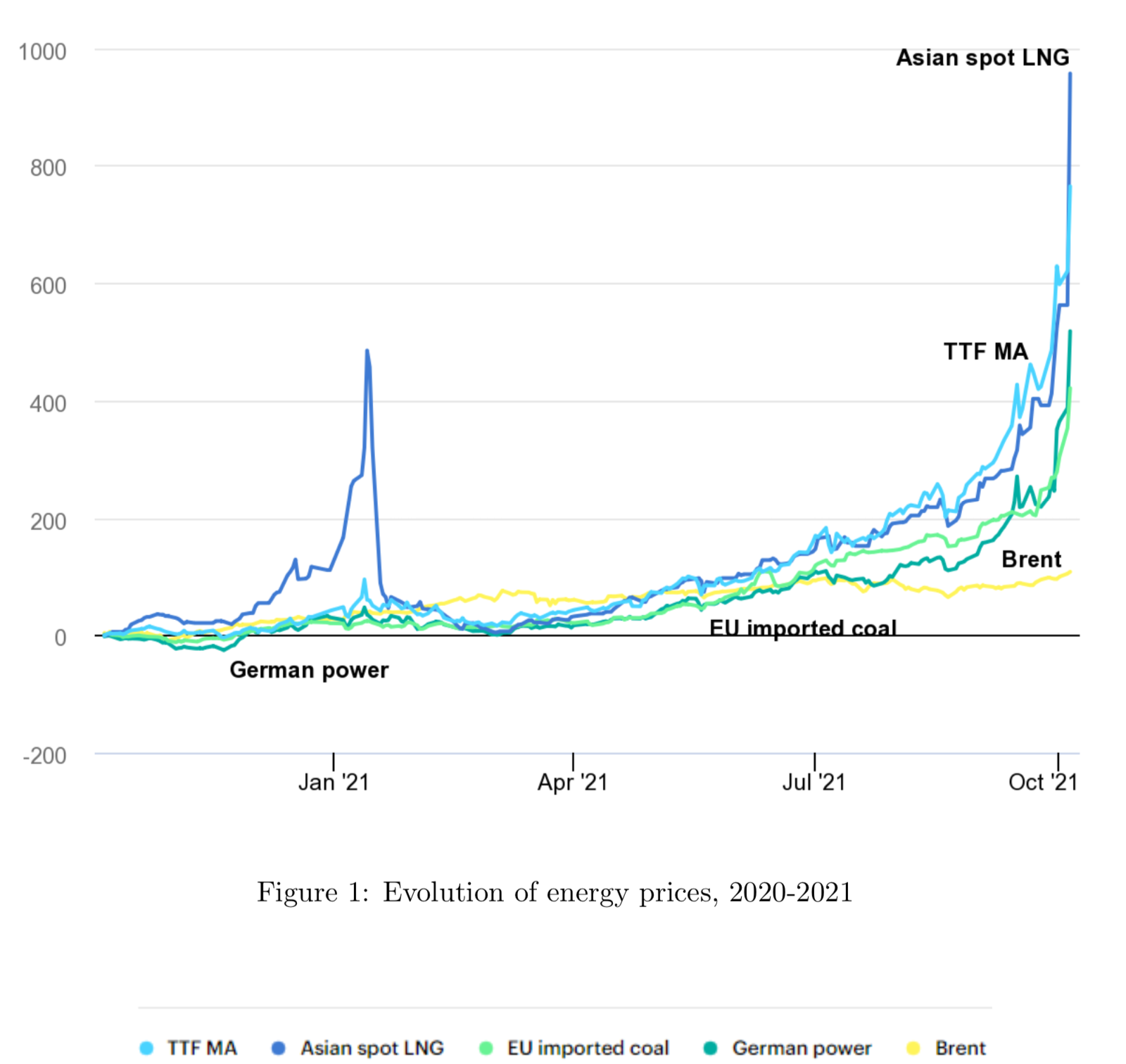}
    \caption{Evolution of energy prices, 2020-2021.}
    \label{fig1}
\end{figure}

The higher gas and coal prices, combined with rising European carbon prices, have resulted in higher electricity prices. In Germany, electricity prices leaped last week to their highest level on record, up more than six times from a year ago \cite{hasanli2019sentiment}.

In India, the economic recovery and related increase in energy demand are causing a coal shortage. India’s domestic coal mining, which accounts for 80\% of the country’s supply, has been unable to keep pace with demand, and higher international prices are making imports uneconomical. Power plants that rely on imported coal have slowed or even halted operations, and some plants that rely on domestic coal are starting to run out. Despite government efforts to address the shortages, several Indian states have suffered serious power shortages in recent days, affecting both residential and industrial customers \cite{alvarez2021behind}.

However Governments is coming up with interim solutions, such as lowering the tax rates and extra levies applied to energy bills, which in some countries can make up half of the final price. The Spanish government has temporarily cut the special electricity tax from 5.1\% to 0.5\% – the minimum under EU law \cite{Heller2022}.

Most empirical studies have used surveys and interviews to measure public opinion, sentiment, awareness and perceptions of renewable energy. The literature finds broad public support for utility products across the United States, Finland, Mexico, Spain, South Korea, Portugal, Greece and worldwide. Surveys and interviews have advantages in gauging individual-level demographic information, such as gender, education, income, distance to energy facilities, and previous experience with renewable energy such as electricity generation, technologies, which is one of the key determinants of individuals’ preferences regarding utility bills \cite{kim2021public}.

However, surveys and interviews are limited in gauging temporal dynamics and geographical variations in public opinions. Researchers are beginning to use social media, especially Twitter, to examine public sentiment toward energy prices rising these days. Here, we compare four different machine learning techniques for sentiment analysis and find that which model will achieve the higher accuracy than others for sentiment classification on electricity prices related tweets.

\section{Twitter Sentiment Analysis Framework}
\label{sec:methodology}

\begin{figure}[ht]
    \centering
    \includegraphics[width=15cm]{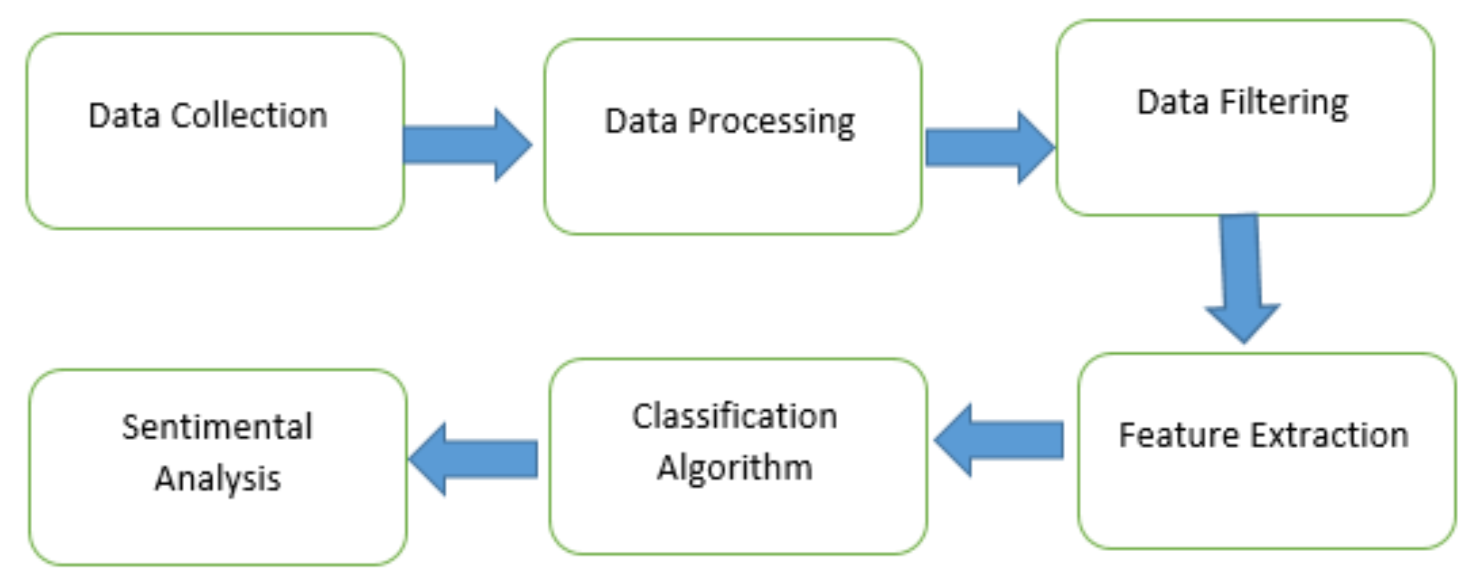}
    \caption{Framework for Twitter sentiment analysis.}
    \label{fig2}
\end{figure}

\subsection{Data collection}
Data collection is the process of extracting tweets from social media, specifically Twitter, which is one of the most popular social media platforms for sharing emotions and opinions. Using API provided by social media platform, ones can collect data in a streaming fashion. An example is Twitter API that allows to extract tweets by hashtags, NewsAPI to extract news by category from different news publishers.

\subsection{Data processing}
Data processing involves tokenization which is the process of splitting the tweets into individual words called tokens. Tokens can be split using whitespace or punctuation characters. It can be unigram or bigram depending on the classification model used. The bag-of-words model is one of the most extensively used model for classification. It is based on the fact of assuming text to be classified as a bag or collection of individual words with no link or interdependence. The simplest way to incorporate this model in our project is by using unigrams as features. It is just a collection of individual words in the text to be classified, so, we split each tweet using whitespace.

\subsection{Data Filtering}
The preprocessing of the data is a very important step as it decides the efficiency of the other steps down in line. It involves syntactical correction of the tweets as desired. The steps involved should aim for making the data more machine readable in order to reduce ambiguity in feature extraction \cite{gupta2017study}. Below are a few steps used for pre-processing of tweets.

\begin{itemize}
    \item  Converting upper case to lower case - In case we are using case sensitive analysis, we might take two occurrence of same words as different due to their sentence case. It important for an effective analysis not to provide such misgivings to the model \cite{gupta2017study}.
    \item Stop word removal - Stop words that don’t affect the meaning of the tweet are removed (for example and, or, still etc.) \cite{gupta2017study}.
    \item Twitter feature removal - User names and URLs are not important from the perspective of future processing, hence their presence is futile. All usernames and URLs are converted to generic tags or removed.
    \item Stemming - Replacing words with their roots, reducing different types of words with similar meanings. This helps in reducing the dimensionality of the feature set.
    \item Special character and digit removal - Digits and special characters don’t convey any sentiment. Sometimes they are mixed with words, hence their removal can help in associating two words that were otherwise considered different \cite{gupta2017study}.

\end{itemize}

\subsection{Feature Extraction}
A feature is a piece of information that can be used as a characteristic which can assist in solving a problem (like prediction). The quality and quantity of features is very important as they are important for the results generated by the selected model \cite{kim2021public}.
Selection of useful words from tweets is feature extraction.
\begin{itemize}
    \item Unigram features – one word is considered at a time and decided whether it is capable of being a feature \cite{gupta2017study}.
    \item n-gram features – more than one word is considered at a time \cite{gupta2017study}.
    \item External lexicon – use of list of words with predefined positive or negative sentiment \cite{gupta2017study}.
\end{itemize}

\subsection{Sentiment Polarity}

Each tweet is then labeled with a sentiment with two possible values: negative or positive. We used a list of English positive and negative opinion words or sentiment words (around 7000 words). There exists a variety of sentiment analysis algorithms able to capture positive and negative sentiment, some specifically designed for short, informal texts. In this work, we first determined the sentiment polarity of each tweet where positive represents the positive words count and negative the negative words count in the tweet. This variable captures well our assumptions about the ordering of the sentiment values and the distances between them. In some cases, the measure polarity fails to capture the degree of emotionality of the tweet because the positive and negative sentiment scores cancel out each other (Sentiment score = 0, although the tweet is actually heavily emotional and not neutral as the measure might indicate). Therefore, we used the following definition:

   \hspace{2cm} \textit{T = 1 (positive tweet) if Sentiment score is greater than or equal to 0.1}
   
   \hspace{2cm} \textit{T = 0 (negative tweet) if Sentiment score is less than equal 0.1} \\

\section{Experimental Setup}

The tools and libraries listed below were used to conduct this research work..

\subsection{Python}

Python is a high level, interpreted programming language which is very popular for its code readability and compact line of codes. Python provides a large standard library which can be used for various applications for example natural language processing, machine learning, data analysis etc. It is favored for complex projects, because of its simplicity, diverse range of features and its dynamic nature \cite{ramadhan2017sentiment}.

\subsection{NLTK}
Natural Language toolkit (NLTK) is a library in python, which provides the base for text processing and classification. Operations such as tokenization, tagging, filtering, text manipulation can be performed with the use of NLTK.

NLTK library is used for creating a bag-of words model, which is a type of unigram model for text. In this model, the number of occurrences of each word is counted. The data acquired can be used for training classifier models. The sentiment of the entire tweets is computed by assigning subjectivity score to each word using a sentiment lexicon.

\subsection{Scikit learn}
The Scikit-learn project started as scikits.learn, a Google Summer Code project by David Cournapeau. It is a powerful library that provides many machine learning classification algorithms, efficient tools for data mining and data analysis. Below are various functions that can be performed using this library \cite{scikit-learn}:
\begin{itemize}
    \item Classification - Identifying the category to which a particular object belongs.
  \item Regression - Predicting a continuous-valued attribute associated with an object.
 \item Clustering - Automatic grouping of similar objects into sets.
 \item Dimension Reduction - Reducing the number of random variables under consideration.
 \item Model selection - Comparing, validating and choosing parameters and models.
 \item Preprocessing - Feature extraction and normalization in order to transform input data for use with machine learning algorithm.
\end{itemize}

In order to work with scikit-learn, it is required to install NumPy on the system.

\subsection{Numpy}

NumPy is the fundamental package for scientific computing with Python. It provides a high- performance multidimensional array object, and tools for working with these arrays. It contains among other things \cite{harris2020array}:

\begin{itemize}
    \item A powerful N-dimensional array object.
    \item Sophisticated (broadcasting) functions
    \item  Tools for integrating C/C++ and Fortran code • Useful linear algebra, Fourier transform, and
    \item Random number capabilities
\end{itemize}

\section{Methodology}

Data used in this study is from one main source - Twitter. Twitter has been a valuable source for opinion mining, but the manual classification of a large number of tweets is difficult and time-consuming. Thus, we use NLP and ML methods to automatically detect public opinion on raised electricity bills.

\begin{figure}[!ht]
    \centering
    \includegraphics[width=15cm]{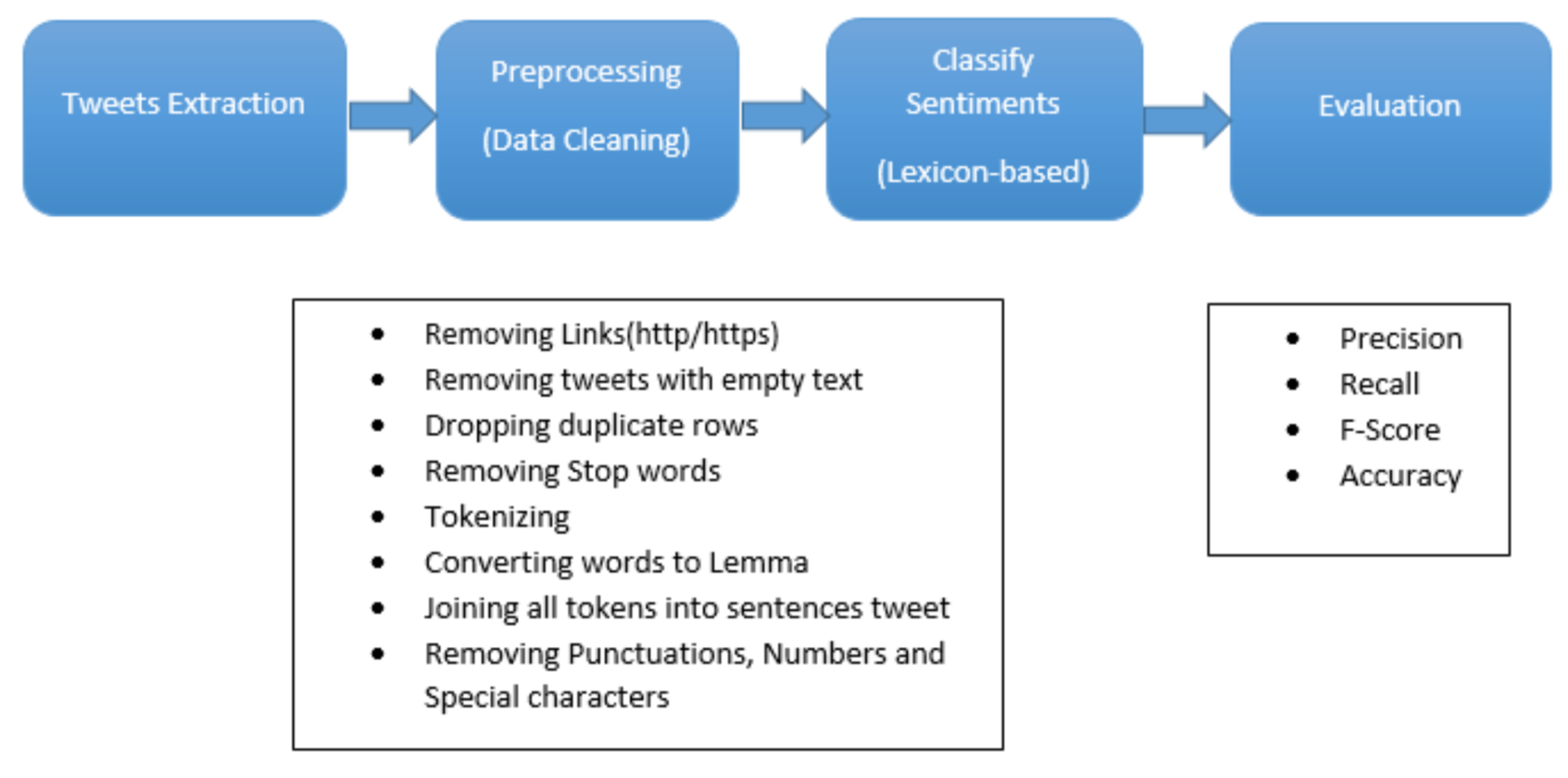}
    \caption{Electricity Tweets Sentiment Analysis Architecture.}
    \label{fig3}
\end{figure}

Figure \ref{fig3} shows the overall sentiment classification architecture of the lexicon based sentiment analysis that consist of Tweets extractions or scraping from Twitter, the preprocessing of data, classifying sentiment using English, and the evaluation of the model using the precision, recall, F-Score and the accuracy.

\subsection{Tweets Extraction}
The initial phase is the data gathering. The Twitter Application Program Interface (API) was used to collect tweets, which are posts created by individuals on Twitter, specific to energy prices. Tweepy executes multiple queries containing trending keywords such as "Electricity tweets", "Electricity prices", "Electricity bill", "Energy bill", and "Energy prices". The dataset comprises attributes like "the tweet’s text," "who sent the tweet," the date the tweet is created. Total of 8731 tweets were collected from Twitter for this project using python programming in Jupyter Notebook from two countries: United Kingdom and India.

\begin{table}[!htb]
    \centering
    \begin{tabular}{|c|c|}
    \hline
        Country& Number of Tweets   \\ \hline
         United Kingdom & 5297   \\\hline 
        India & 3434 \\ \hline 
    \end{tabular}
    \caption{No. of tweets gathered}
    \label{tab:No of tweets}
\end{table}

Table \ref{tab:No of tweets} shows the number of tweets for each country that includes UK and India tweets regarding electricity prices, with 5297 and 3434 tweets for each country respectively. It is quite hard to extract data from twitter because of the Twitter changes its API terms from time to time and you can only extract tweets for the last 7 days and you are only allowed to scrape Twitter tweets every 15 minutes interval depending on your authentication method. We can use data augmentation techniques to further generate more tweets globally \cite{imran2022impact, wei2019eda, shorten2021text}, or to generate class-specific tweets \cite{fatima2022systematic}.

\begin{figure}[ht]
    \centering
    \includegraphics[width=16cm]{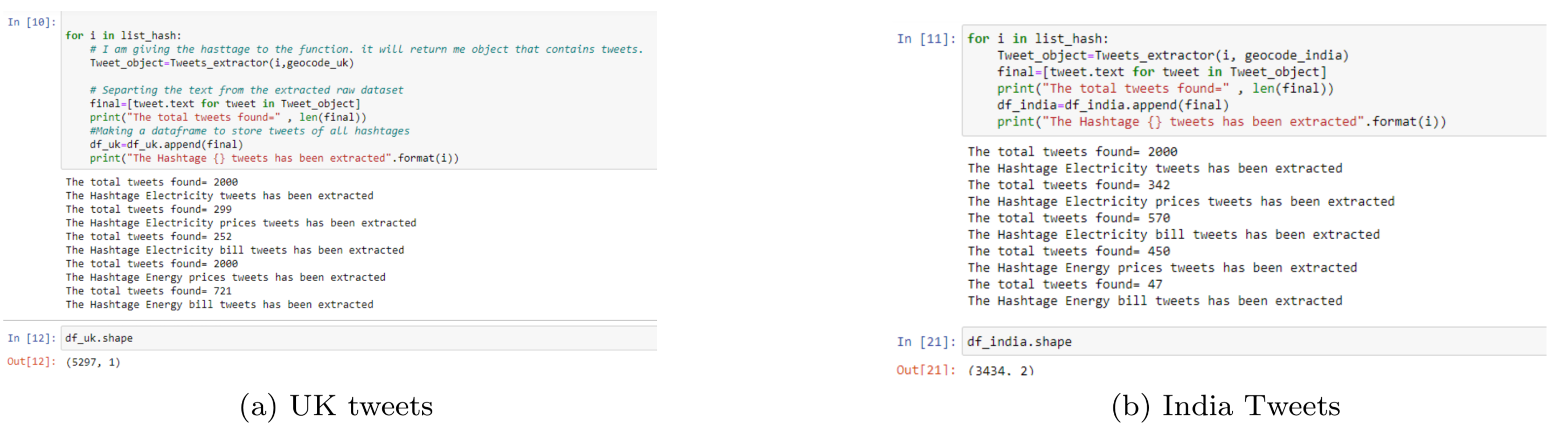}
    \caption{UK and India Tweets.}
    \label{fig4}
\end{figure}

We restricted the tweets limit to 2000 on each hashtag which we used in sentiment analysis. The tweets which contain text, hyperlinks, different symbols, emoticons and short words like lol (laughing out loud) that makes a great challenge when working with Twitter data. Here are the samples of the tweets we have extracted from Twitter using the above mentioned keywords from both countries:

\begin{figure}[ht]
    \centering
    \includegraphics[width=10cm]{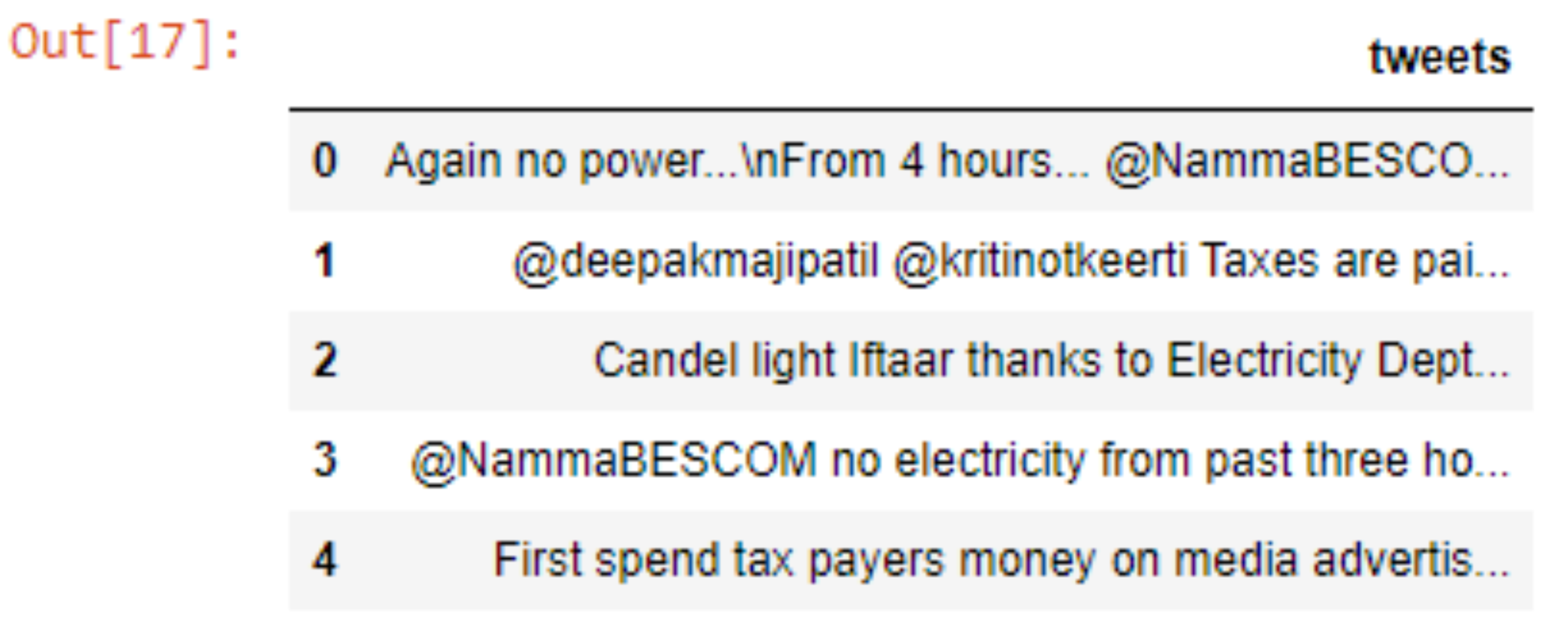}
    \caption{Extracted tweets.}
    \label{fig5}
\end{figure}

\subsection{Text Preprocessing}

The cleaning is quite complicated in Twitter tweets but in this project, we use some regular expression techniques to clean our data. We have used \textit{findall()} function finds \textit{*all*} the matches and returns them as a list of strings, with each string representing one match.

\begin{figure}[ht]
    \centering
    \includegraphics[width=10cm]{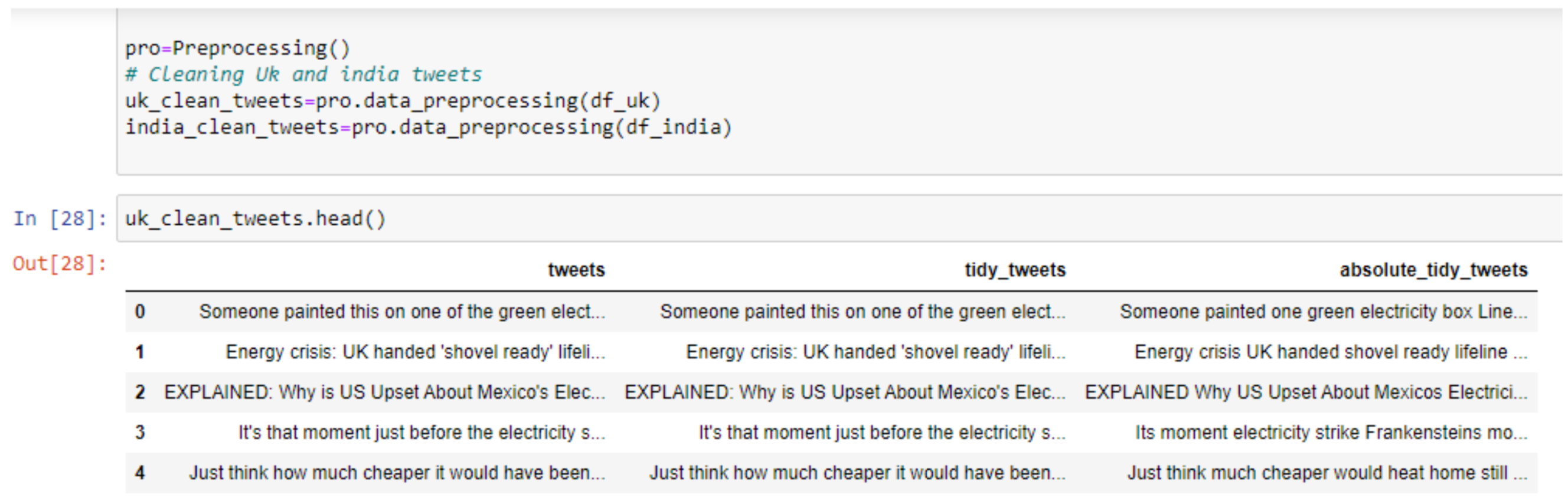}
    \caption{Cleaned Tweets for UK and India.}
    \label{fig6}
\end{figure}

\subsection{Classify Sentiments}
After the cleaning process, we can do the sentiment analysis. NLTK’s Sentiment Intensity Analyzer (VADER) is used to know the sentiment type of every tweet. This is the lexicon based approach.
Here, we used the function \textit{sid.polarityscores(text)} that takes text as an input and return the polarity score such as positive and negative.

\begin{figure}[ht]
    \centering
    \includegraphics[width=10cm]{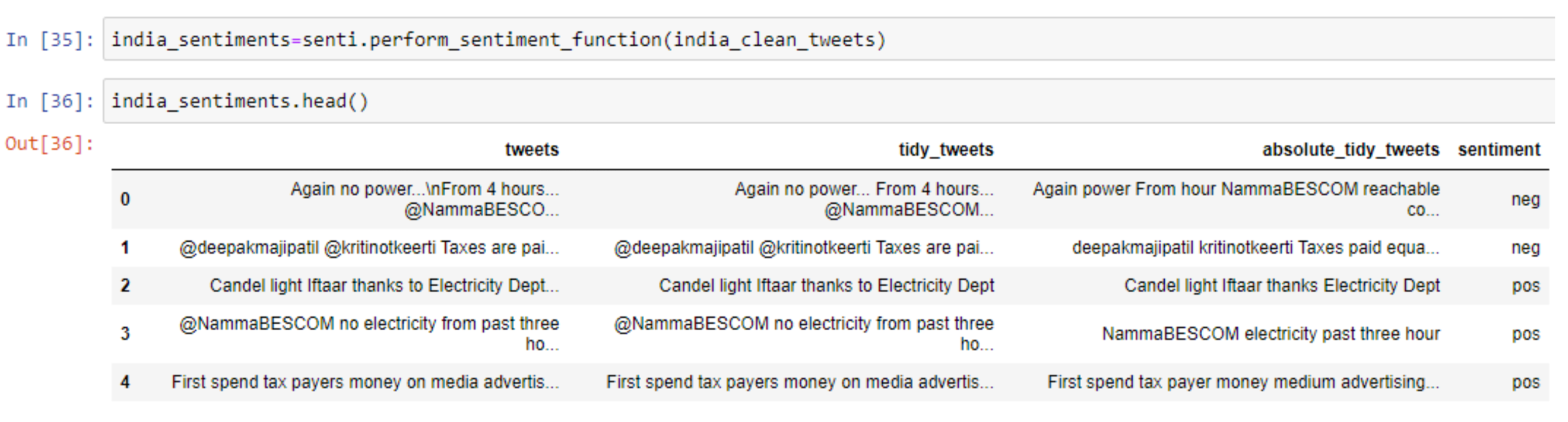}
    \caption{Sentiment Analysis with two classes: Positive and Negative.}
    \label{fig7}
\end{figure}

Further, we saved the data in separate csv files for the UK and India with the polarity scores.

\subsection{Story Generation and Visualization}

Next, the \textit{csv} file is read using python for both countries for the visualization of most common text posted on twitter regarding electricity prices. WordCloud was used to generate a graphical representation of the most frequently appearing words in the tweets such as positive and negative for both selected countries. WordCloud basically gave summary of isolated words without knowing their linguistic meaning or relations.

\begin{figure}[ht]
    \centering
    \includegraphics[width=16cm]{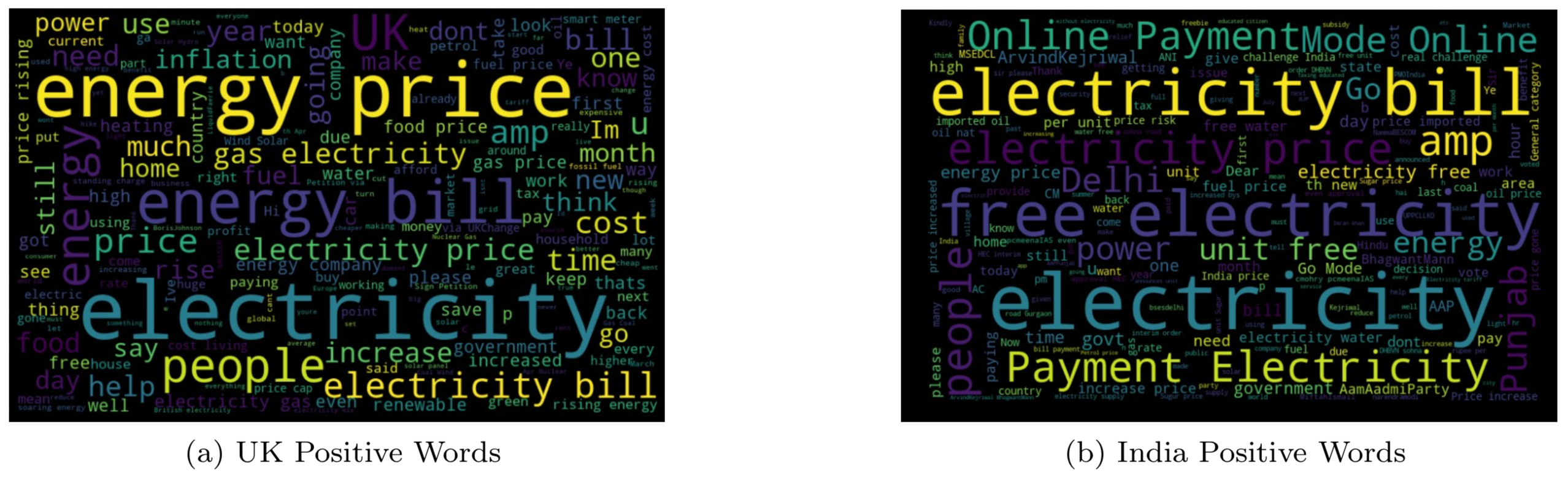}
    \caption{ WordCloud of positive tweeets for UK and India.}
    \label{fig8}
\end{figure}

\begin{figure}[ht]
    \centering
    \includegraphics[width=16cm]{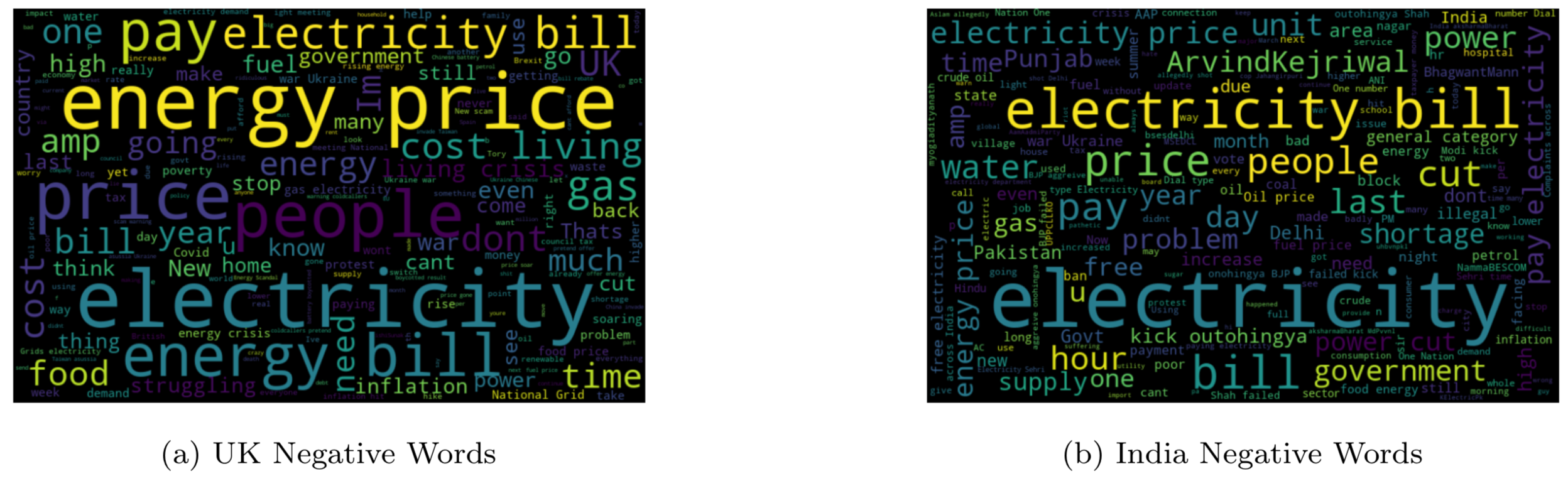}
    \caption{ WordCloud of negative tweeets for UK and India.}
    \label{fig9}
\end{figure}

In Figures \ref{fig8} and \ref{fig9} 1959 tweets are classified as positive sentiment, and 661 tweets are classified as negative sentiment.

\subsection{Feature Extraction}

We need to convert textual representation in the form on numeric features. We have two popular techniques to perform feature extraction:

\begin{itemize}
    \item TF-IDF (Term Frequency - Inverse Document Frequency)
    \item N-Gram
\end{itemize}

In this study, we have used The term frequency- inverse document frequency (TF-IDF) which is a well-recognized method to evaluate the importance of a word in a document. Term Frequency of a particular term (t) is calculated as number of times a term occurs in a document to the total number of words in the document. IDF (Inverse Document Frequency) is used to calculate the importance of a term. There are some terms like “is”, “an”, “and” etc. which occurs frequently but don’t have importance. IDF is calculated as IDF (t) = log(N/DF), where \textit{N} is the number of documents and \textit{DF} is the number of document containing term \textit{t}. TF-IDF is a better way to convert the textual representation of information into a Vector Space Model (VSM). Suppose there is a document which contains 200 words and out of these 200 words, \textit{like} appears 10 times than term frequency will be 10/250=0.04 and suppose there are 50000 documents and out of \textit{like}. Than IDF (mouse) = 50000/500=100, and TF-IDF (like) will be 0.04*100=4.

We were limited to only bag of words. These features can further be incorporated with semantics \cite{kastrati2016semcon, lu2006enhancing, kastrati2015semcon} and vector space representation models \cite{kastrati2019performance, parraguez2010construction, guo2020nonlinear, kastrati2015improved, erk2012vector} to improve classification performance. 

\subsection{Model Building}
Positive sentiments are labelled as 1 and negative sentiments are labelled as 0.

To validate the accuracy of the system, we used k-fold method and to measure the validity of the tweet confusion matrix is used. In k-fold cross validation, the initial data is randomly partitioned into a subset (fold), each of the same size. The training and testing process is done as many times as k. Figure 10 shows the confusion matrix which is used to assist in calculation of the evaluation system \cite{kim2021public}.

Confusion Matrix is the visual representation of the Actual VS Predicted values. It measures the performance of our Machine Learning classification model and looks like a table-like structure \cite{alvarez2021behind}.

Elements of Confusion Matrix - It represents the different combinations of Actual VS Predicted values. Let’s define them one by one.

\begin{itemize}
    \item TP: True Positive - The values which were actually positive and were predicted positive.

     \item FP: False Positive - The values which were actually negative but falsely predicted as positive. Also known as Type I Error.

    \item FN: False Negative - The values which were actually positive but falsely predicted as negative. Also known as Type II Error.

     \item TN: True Negative - The values which were actually negative and were predicted negative. Next, once we know these values, we can calculate the precision and recall.
\end{itemize}

\begin{figure}[ht]
    \centering
    \includegraphics[width=6cm]{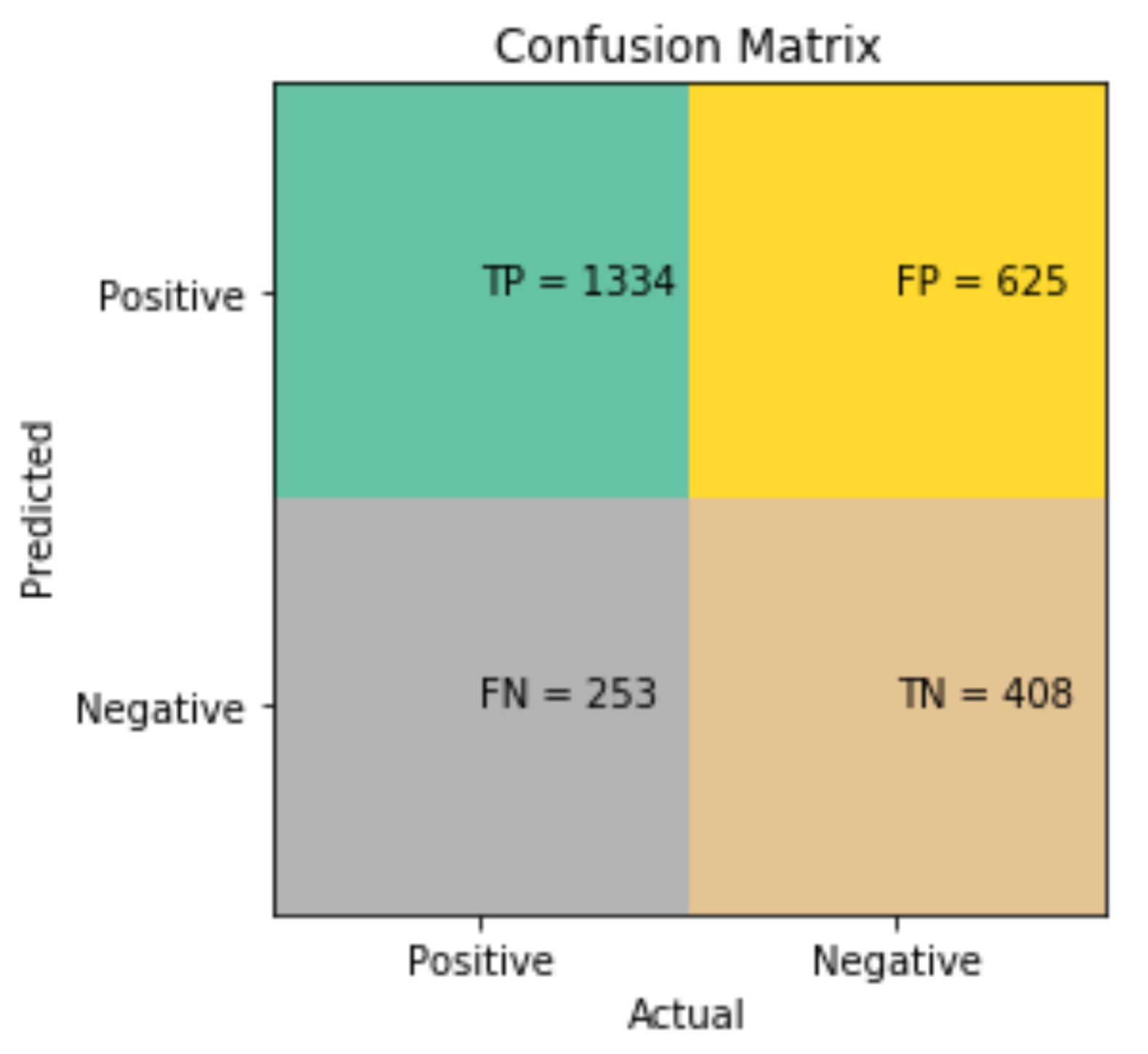}
    \caption{Sentiment Analysis Confusion Matrix.}
    \label{fig10}
\end{figure}

The metric that can be derived from the confusion matrix configuration:

\begin{itemize}
    \item Accuracy - The value of the entire true predicted against all predicted. The formula to obtain accuracy can be seen in equation.
\end{itemize}
    
    \begin{equation}
    \centering
        Accuracy = \frac{TN + TP}{TN + TP + FP + FN}
    \end{equation}

\begin{itemize}
    \item Precision - The value of true positive prediction against all positive prediction. The formula to obtain precision be seen in equation.
\end{itemize}   
    \begin{equation}
    \centering
        Precision = \frac{TP}{TP + FN}
    \end{equation}

\begin{itemize}
    \item Recall - The value of true positive prediction against all actual positive. The formula recall can be seen in equation.
 \end{itemize}   
    \begin{equation}
    \centering
        Recall = \frac{TP}{TP + FP}
    \end{equation}

\begin{table}[!htb]
    \centering
    \begin{tabular}{|c|c|c|c|c|}
    \hline
        Class & Precision & Recall & F1-score & Support   \\ \hline
         1 & 0.87 & 0.91 & 0.89 & 1959   \\\hline 
         0 & 0.69 & 0.61 & 0.65 & 661   \\\hline 
         Macro avg & 0.78 & 0.76 & 0.77 & 2620 \\\hline 
         Weighted avg & 0.83 & 0.83 & 0.83 & 2620 \\ \hline 
    \end{tabular}
    \caption{Classification report}
    \label{tab: Classification report}
\end{table}

\section{Classifier}

Majority of the algorithms for sentiment analysis are built from a classifier that was trained using a collection of annotated text data. In this section, we presented the four classifiers used in this study for comparing the accuracy of each model.

\subsection{Naive Bayes Classifier}
Naiıve Bayes (NB) classifier is a probabilistic classifier which is created based on Bayes theorem but with strong assumptions regarding independence. It is used because of its simplicity aside from the fact that it is commonly used for sentimental analysis application by predicting the words that belong to a class \cite{contreras2018lexicon}.

\begin{figure}[ht]
    \centering
    \includegraphics[width=6cm]{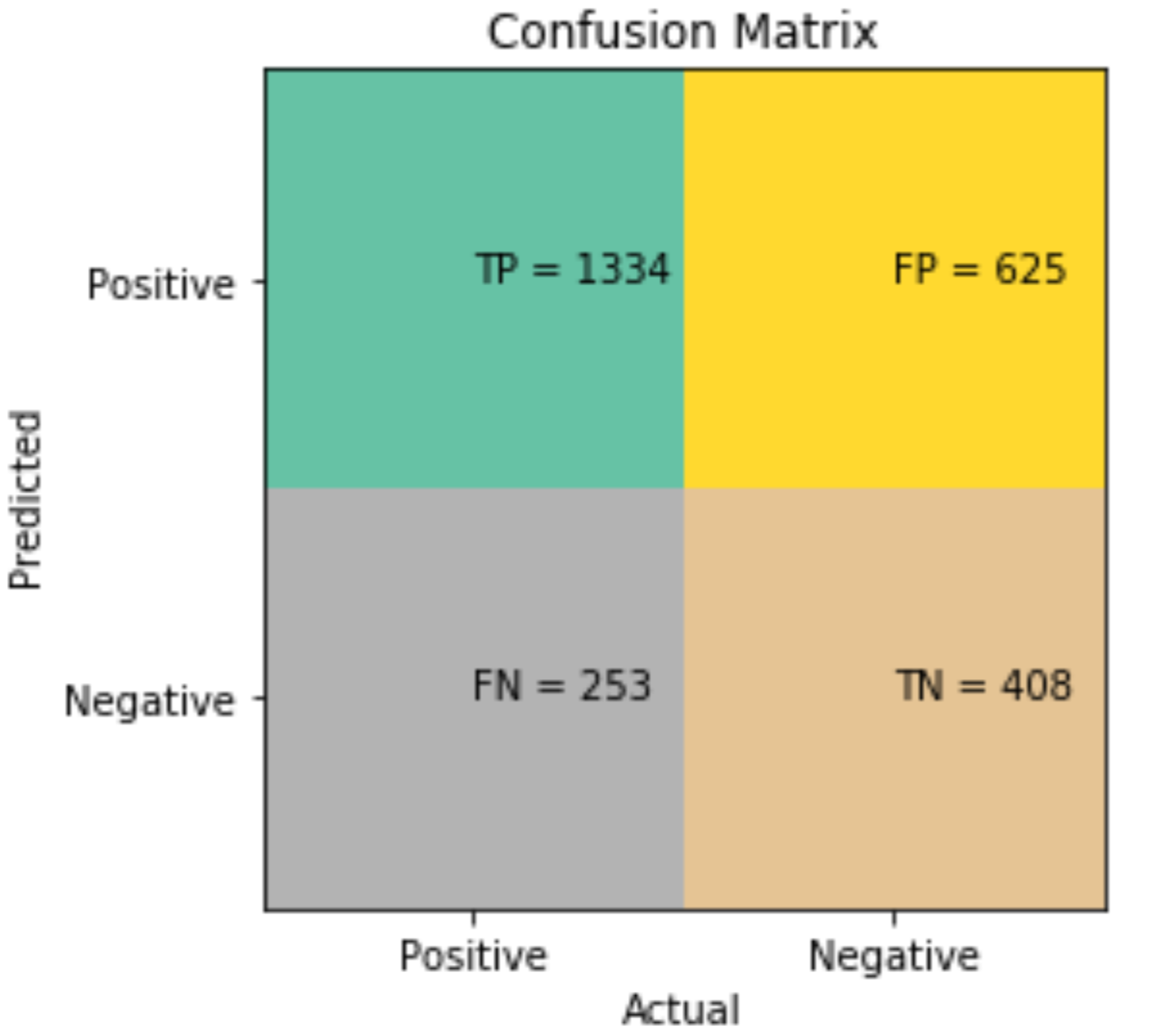}
    \caption{Naive Bayes confusion matrix.}
    \label{fig11}
\end{figure}

\begin{figure}[ht]
    \centering
    \includegraphics[width=8cm]{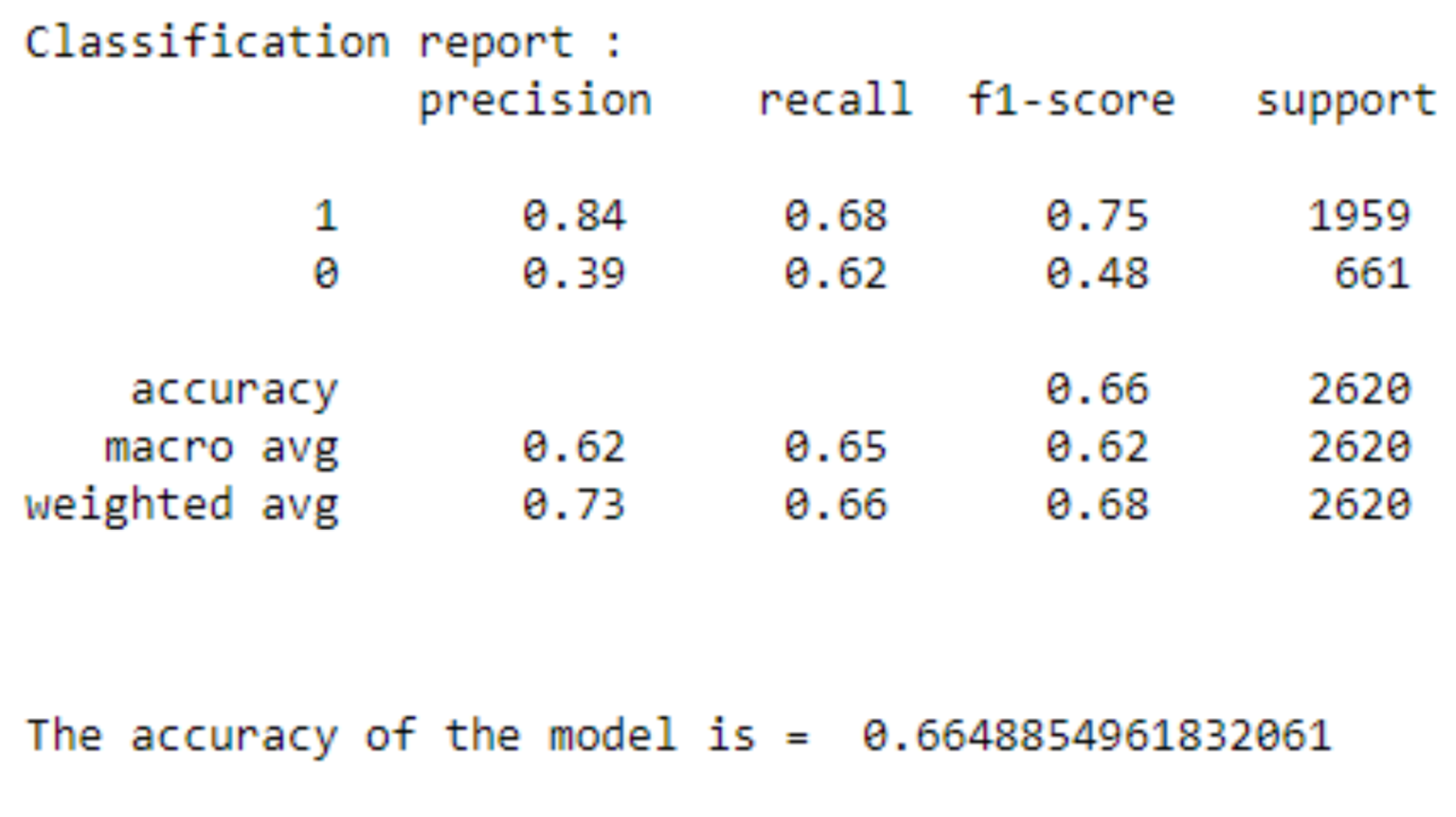}
    \caption{Naive Bayes classification report.}
    \label{fig12}
\end{figure}

\subsection{Decision Tree Classifier}

Decision Tree (DT) classifier delivers a hierarchical decomposition of the training data space where the condition of the attribute value is applied to for data division. It repetitively divides the working is plot into subpart by identifying lines. It is a flowchart-like structure that is consists of an internal node that signifies a test on an attribute and branch that signifies the test results and having the class labels or class distribution indicates the leaf nodes \cite{contreras2018lexicon}. It is a supervised classifier model which uses data with known labels to form the decision tree and then the model is applied to the test data.

\begin{figure}[ht]
    \centering
    \includegraphics[width=6cm]{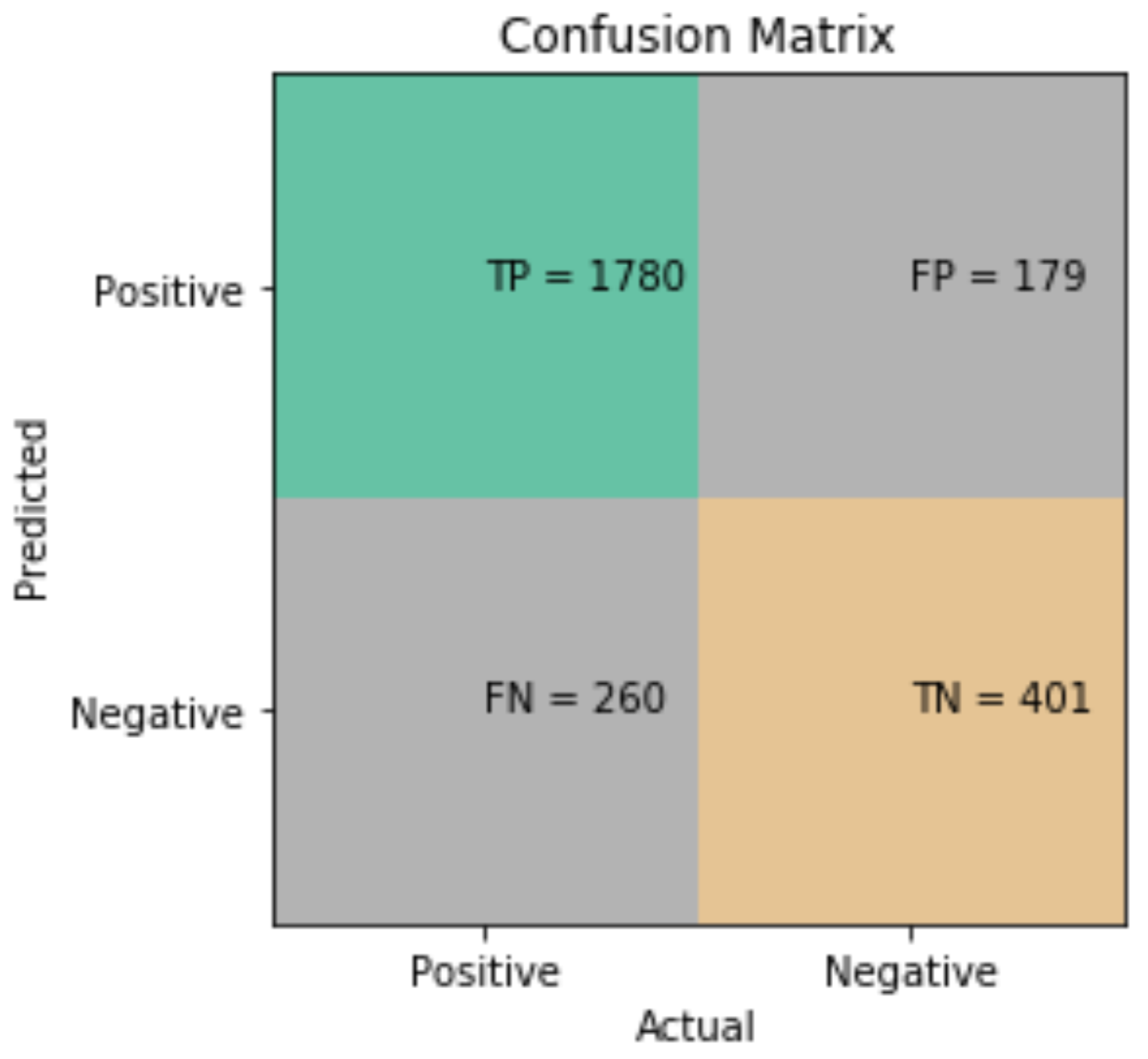}
    \caption{Decision Tree confusion matrix.}
    \label{fig13}
\end{figure}

\begin{figure}[ht]
    \centering
    \includegraphics[width=8cm]{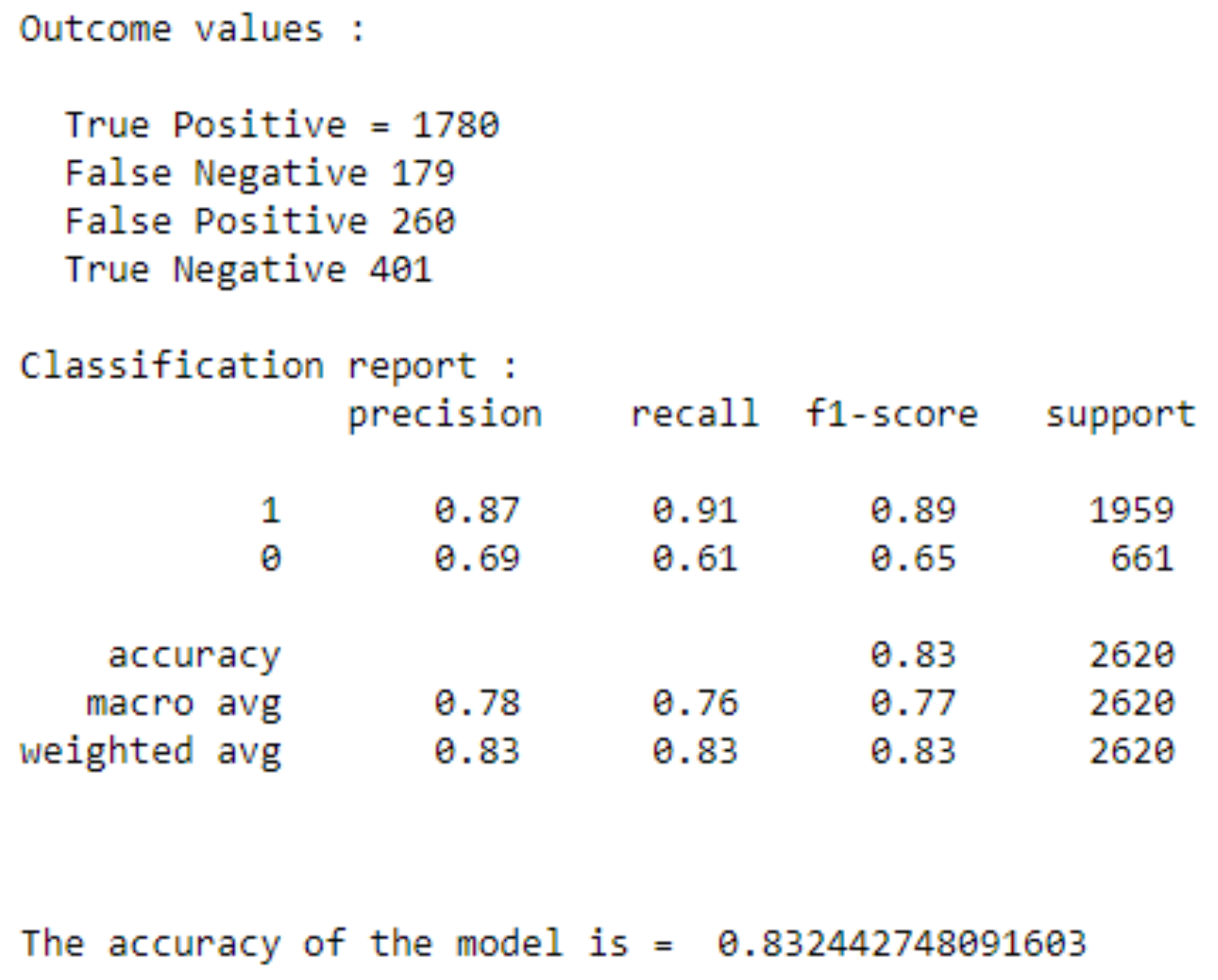}
    \caption{Decision Tree classification report.}
    \label{fig14}
\end{figure}

\subsection{Random Forest Classifier}
Random forest, which were formally proposed in 2001 by Leo Breiman and Adele Cutler, are part of the automatic learning techniques. This algorithm combines the concepts of random subspaces and ”bagging”. The decision tree forest algorithm trains on multiple decision trees driven on slightly different subsets of data \cite{al2018random}. In random forest classification method, many classifiers are generated from smaller subsets of the input data and later their individual results are aggregated based on a voting mechanism to generate the desired output of the input data set \cite{al2018random}.

\begin{figure}[ht]
    \centering
    \includegraphics[width=6cm]{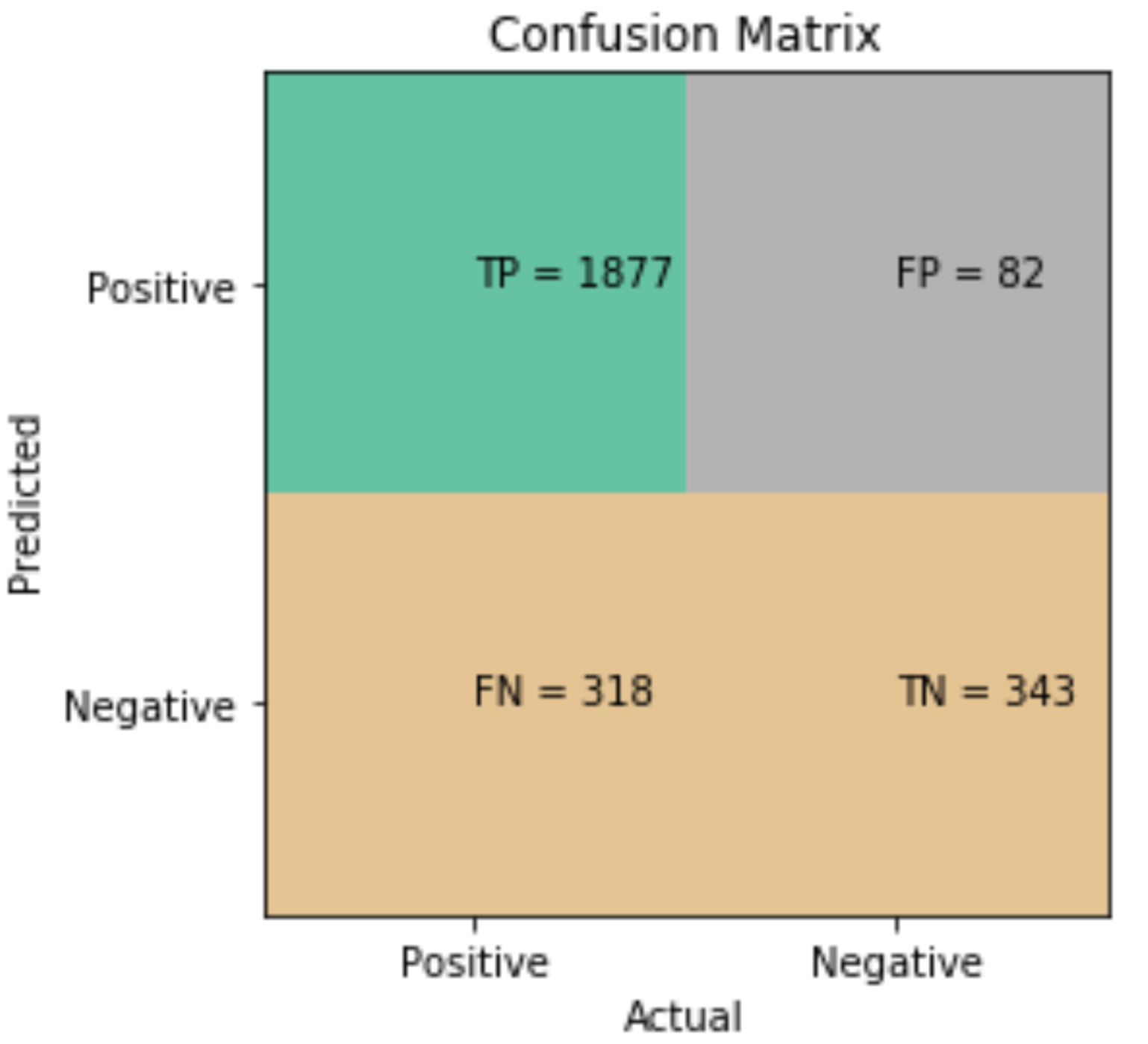}
    \caption{Random Forest confusion matrix.}
    \label{fig15}
\end{figure}

\begin{figure}[ht]
    \centering
    \includegraphics[width=8cm]{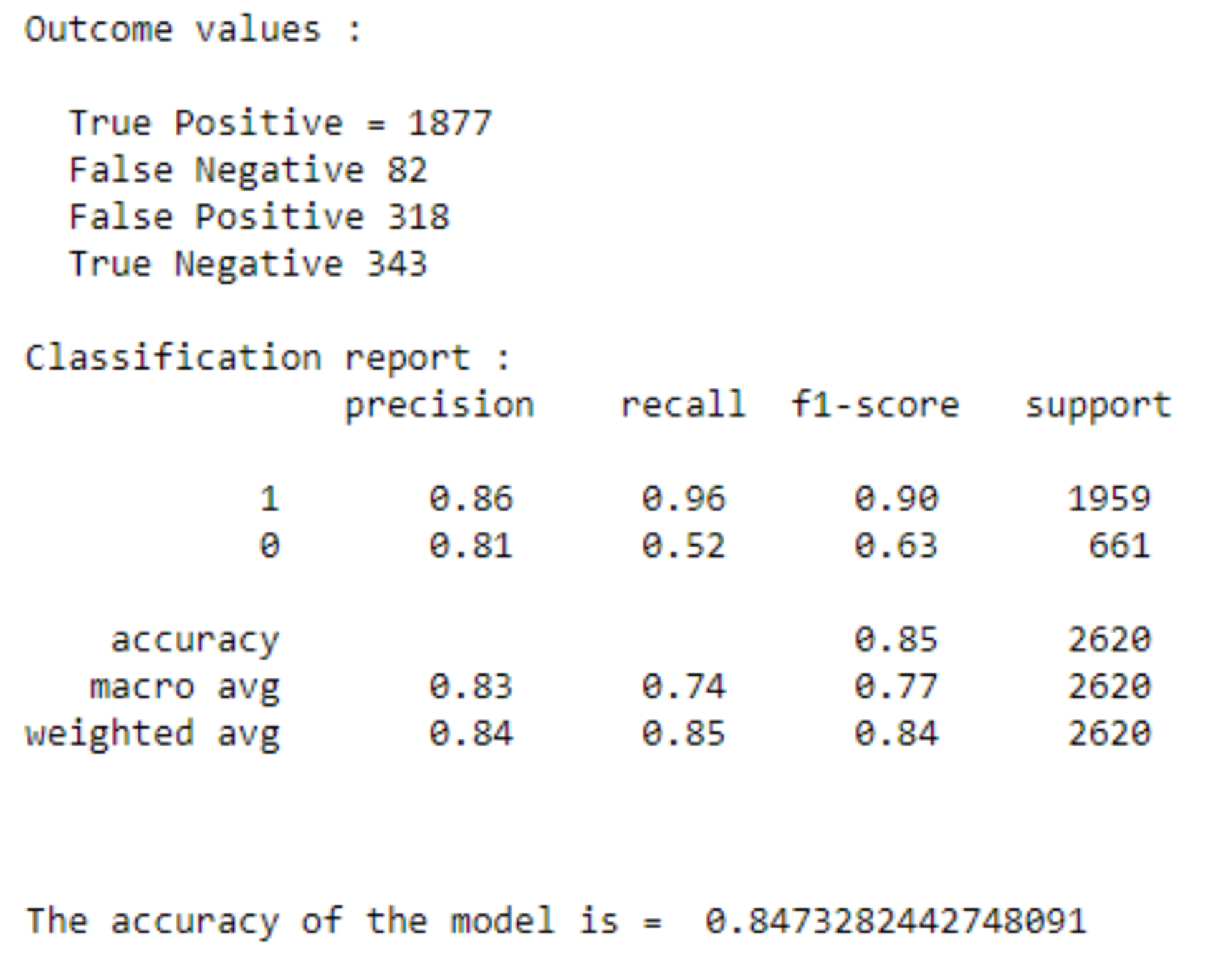}
    \caption{Random Forest classification report.}
    \label{fig16}
\end{figure}

\subsection{Logistic Binary Classifier}

Logistic Regression is a Machine Learning classification algorithm that is used to predict the probability of a categorical dependent variable \cite{hasanli2019sentiment}. It is used to determine the output or result when there are one or more than one independent variables. The output value can be in form of 0 or 1 i.e. in binary form \cite{tyagi2018sentiment}. This method is a generalized linear regression method for learning a mapping from any number of numeric variables to a binary or probabilistic variable \cite{hasanli2019sentiment}.

\begin{figure}[ht]
    \centering
    \includegraphics[width=6cm]{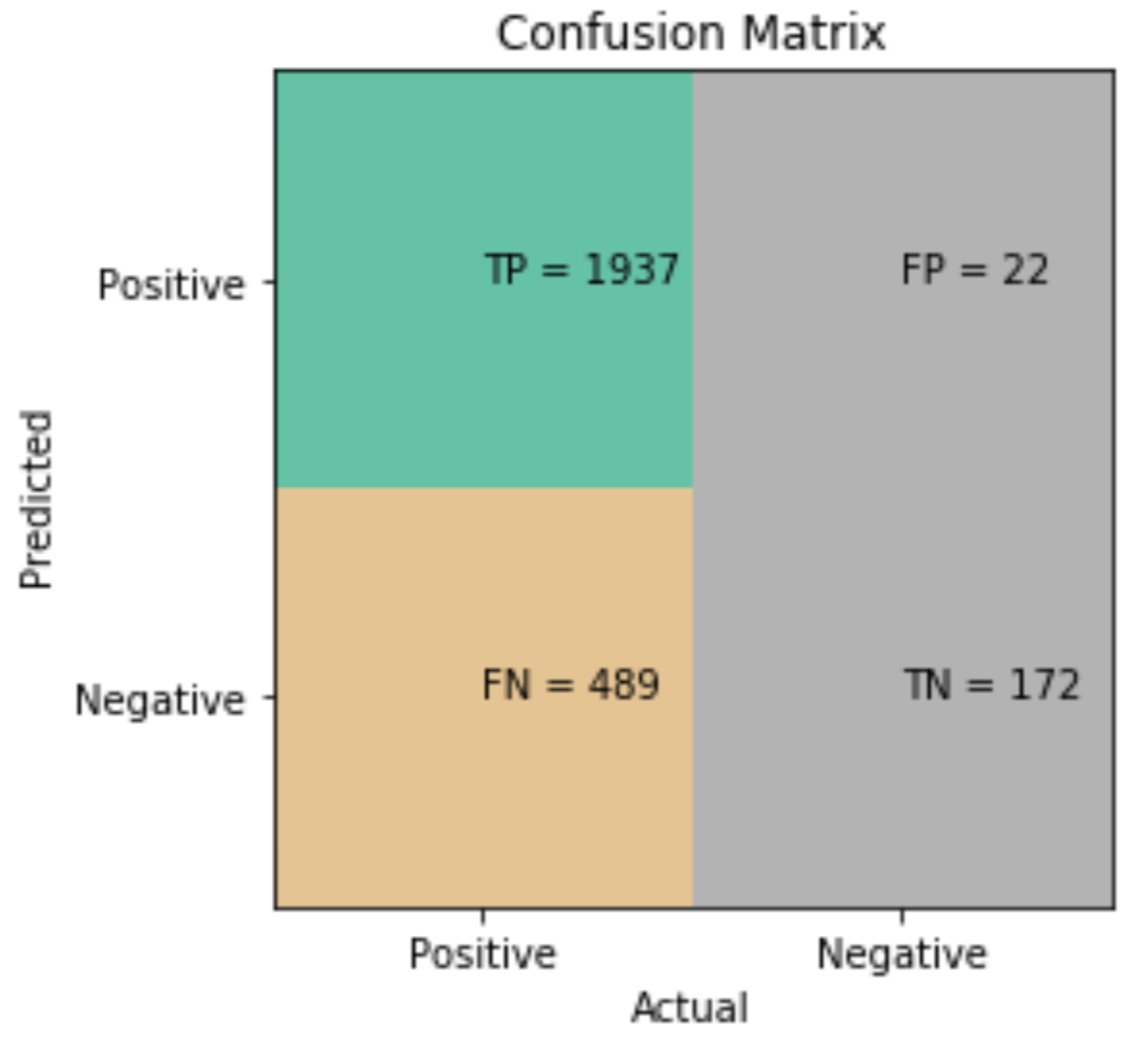}
    \caption{Logistic Binary confusion matrix.}
    \label{fig17}
\end{figure}

\begin{figure}[ht]
    \centering
    \includegraphics[width=8cm]{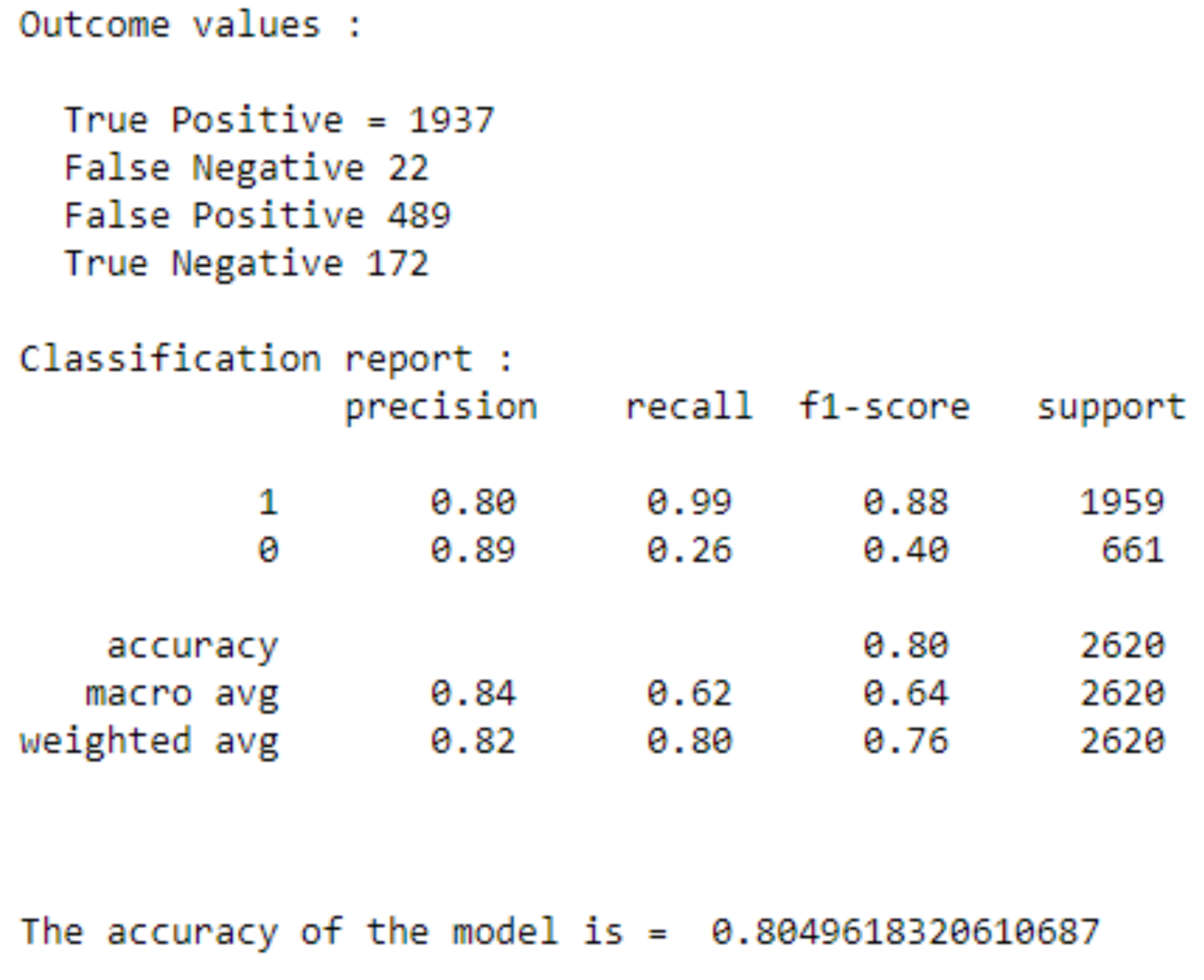}
    \caption{Logistic Regression classification report.}
    \label{fig18}
\end{figure}

\subsection{ROC curve}

AUC (Area Under The Curve) - ROC curve is a performance measurement for the classification problems at various threshold settings. ROC is a probability curve and AUC represents the degree or measure of separability \cite{ruz2020sentiment}.

An excellent model has AUC near to the 1 which means it has a good measure of separability. A poor model has an AUC near 0 which means it has the worst measure of separability. In fact, it means it is reciprocating the result. It is predicting 0s as 1s and 1s as 0s. And when AUC is 0.5, it means the model has no class separation capacity whatsoever.By analogy, the Higher the AUC, the better the model is at distinguishing \cite{imran2020cross}.
When AUC is 0.7, it means there is a 70\% chance that the model will be able to distinguish between positive class and negative class such as we can see in Figure 15 that decision tree and Random Forest are good in classification \cite{ruz2020sentiment}.

\begin{figure}[ht]
    \centering
    \includegraphics[width=12cm]{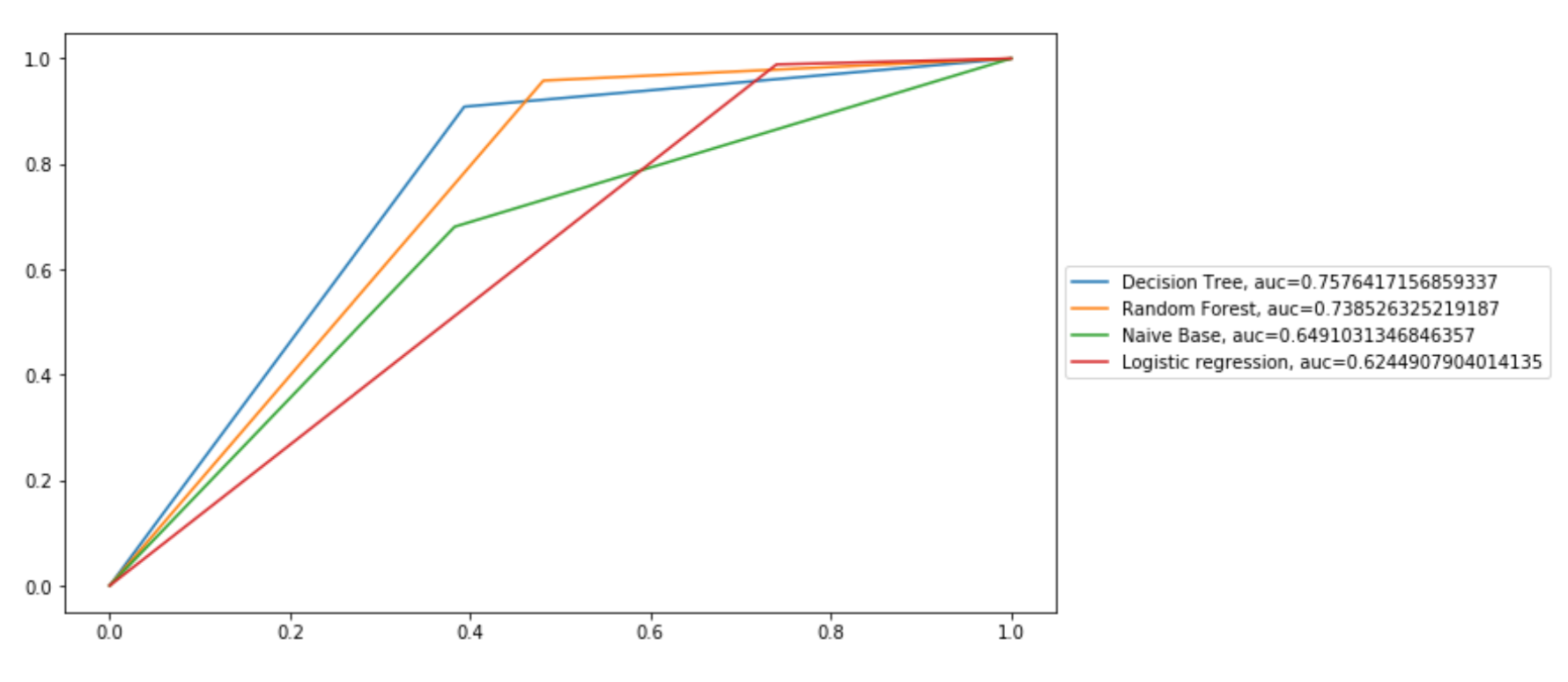}
    \caption{ROC Curve for all models.}
    \label{fig19}
\end{figure}

The worst case situation is when AUC is approximately 0.5, the model has no discrimination capacity to distinguish between positive class and negative class. However all the classifier used in this research work are good to classify the polarity of data that is negative and positive.

\section{Results}
Table \ref{tab: Accuracy} shows the accuracy and runtime score of Dataset where Random forest and Decision Tree got the almost same accuracy value i.e 0.84 and 0.83 with a little difference which means both classifiers can be used to attain a higher percentage of accuracy while the Naive Bayes is only 0.66 and Logistic Regression 0.80

\begin{table}[!htb]
    \centering
    \begin{tabular}{|c|c|}
    \hline
        Model & Accuracy   \\ \hline
        Naive Bayes & 0.66   \\\hline 
        Decision Tree & 0.83 \\ \hline 
        Random Forest & 0.84  \\ \hline 
        Logistic Regression & 0.80 \\ \hline 
    \end{tabular}
    \caption{Accuracy obtained from different models}
    \label{tab: Accuracy}
\end{table}

Figure 20 displays classification report for all models, where 1 represents positive sentiment and 0 represents negative sentiment. It is clearly seen in the figure that Decision Tree and Random forest has the highest precision score whereas Random Forest and Logistic Regression has the highest Recall value.

\begin{figure}[ht]
    \centering
    \includegraphics[width=8cm]{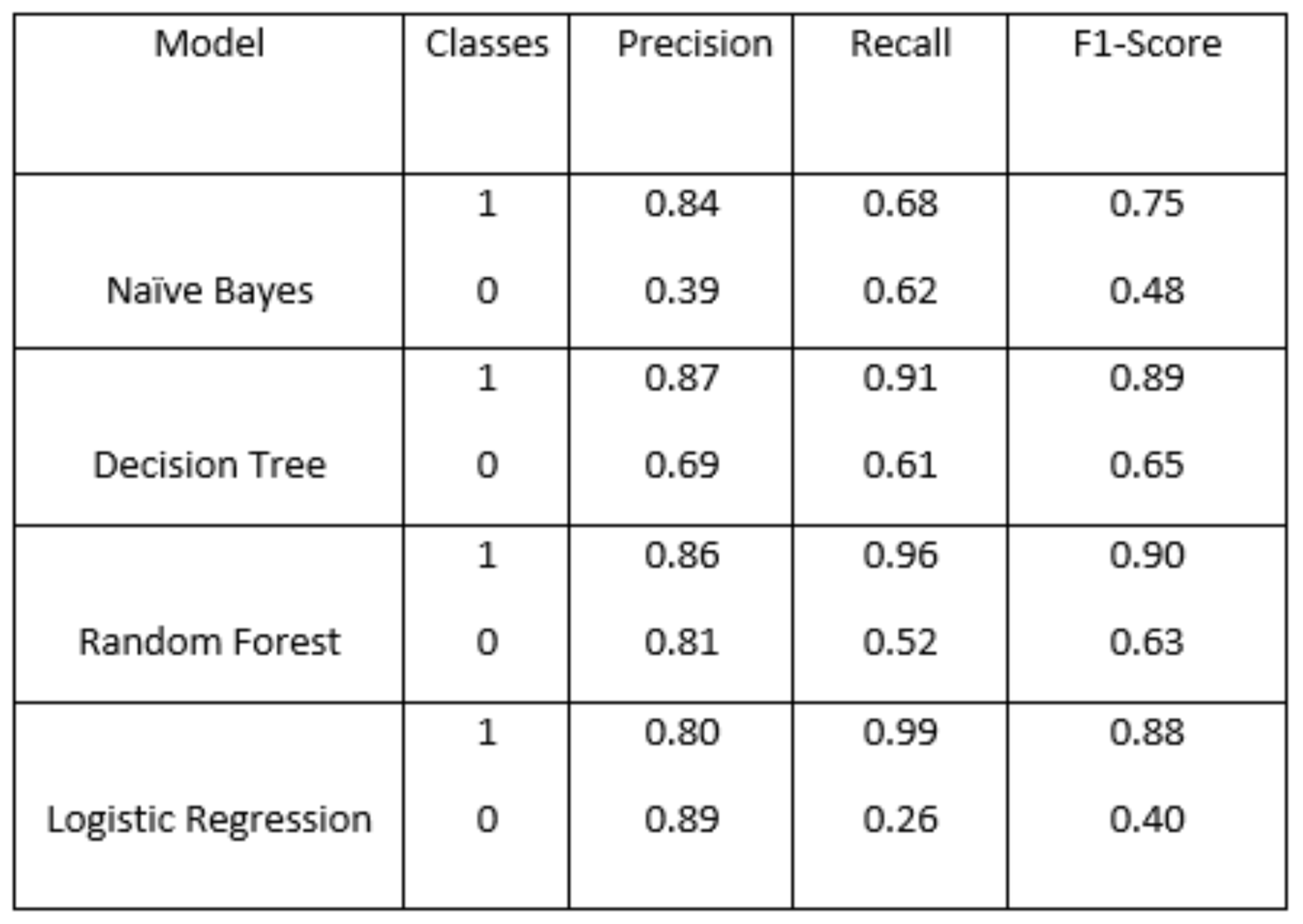}
    \caption{Classification report for all models.}
    \label{fig20}
\end{figure}

\section{Limitations}
Although we explored many machine learning models on analysing sentiments on the twitter data for electricity price hikes, this study is limited due to the small number of available tweets. We only extracted tweets in English. If we are to expand the search to other regions and to extract tweets in other languages for sentiment analysis \cite{chandio2022attention, arora2013sentiment, pandey2015framework}, or use multilingual translation approach \cite{ghafoor2021impact}, then we can increase more tweets. 

We may further be able to improve upon the accuracy by balancing the dataset. Though, due limited number of samples, we didn't train and employ deep learning models, however, combining data augmentation techniques \cite{fatima2022systematic, liu2020survey, kobayashi2018contextual} with data balancing \cite{shaikh2021towards, tepper2020balancing, son2021bcgan}, we can generate many more synthetic samples. 

We have also only used bag of words. In latest techniques, word embeddings are widely used and employed \cite{kastrati2020wet}. With embedding-based models and a large corpora, we can easily employee deep learning models.

\section{Conclusion}
In this study, we discussed the sentiment analysis of public opinions towards hike in electricity prices, using tweet data obtained from social media Twitter. In addition, we performed the text processing from data obtained and used Naive Bayes, Decision Tree, Logistic Binary and Random Forest method to predict the class. From the results of our experiments, it can be seen that the Random Forest model has a better accuracy level (i.e. 0.84) compared to using other methods, such as Naive Bayes which only has an accuracy rate of 0.66 . Other models Decision tree and Logistic Regression has good accuracy level to determine the negative sentient and positive sentiment achieving an accuracy of 0.83 and 0.80, respectively. For future work, we plan to analyze sentiment by incorporating the more complex models. Further, would like to use advanced embedding models, take into account the semantics for improving the results. Apart from data augmentation techniques, we will also include data from another social media, such as Facebook and Instagram.

\bibliography{access.bib}{}
\bibliographystyle{IEEEtran}

\end{document}